\pdfoutput=1
\documentclass[fleqn,10pt]{wlpeerj}

\usepackage[authoryear,round]{natbib}
\usepackage{hyperref}
\usepackage{orcidlink}
\definecolor{color2}{RGB}{255,255,255}
\hypersetup{colorlinks=true,linkcolor=black,citecolor=black,urlcolor=blue!55!black}
\setcitestyle{authoryear,round,aysep={},yysep={;}}
\newcommand{\holmM}{30}
\newcommand{\holmMain}{20}
\newcommand{\holmCL}{10}
\newcommand{\goldFailN}{5}
\newcommand{\goldFailIDs}{518, 596, 646, 701, 1014}
\newcommand{\goldMaxEX}{99.67}
\newcommand{\TabEXBody}{%
Self-hosted, & Qwen2.5-Coder-7B & 42.63 & 46.54 & 43.48 & 45.37 & -- & -- \\
16-bit & Qwen2.5-Coder-14B & 52.02 & 53.52 & 49.87 & 52.09 & -- & -- \\
 & Qwen2.5-Coder-32B & 54.89 & 56.19 & 53.72 & 55.41 & -- & -- \\
 & CodeLlama-7B & 23.73 & 27.12 & 25.42 & 28.16 & 22.62 & -- \\
 & CodeLlama-13B & 26.27 & 30.64 & 26.34 & 29.92 & 25.68 & -- \\
\midrule
API, FP8 & Llama-3.1-8B & 37.09 & 41.07 & 34.75 & -- & -- & -- \\
 & Llama-3.3-70B & 54.69 & 55.67 & 50.46 & -- & -- & -- \\
\midrule
API, OpenRouter & Qwen2.5-Coder-32B & 54.50 & 55.87 & 53.46 & -- & 54.11 & 55.35 \\}
\newcommand{\TabSigBody}{%
Qwen-7B & base $\to$ + corr. & $+3.91$ & 67/7 & $2.1\times10^{-13}$ & $5.9\times10^{-12}$ & $[1.29, 8.17]$ \\
Qwen-14B & base $\to$ + corr. & $+1.50$ & 26/3 & $1.5\times10^{-5}$ & $3.2\times10^{-4}$ & $[0.94, 2.10]$ \\
Qwen-32B & base $\to$ + corr. & $+1.30$ & 27/7 & $8.2\times10^{-4}$ & 0.014 & $[0.45, 2.06]$ \\
Llama-8B & base $\to$ + corr. & $+3.98$ & 71/10 & $1.8\times10^{-12}$ & $4.7\times10^{-11}$ & $[2.77, 5.25]$ \\
Llama-70B & base $\to$ + corr. & $+0.98$ & 44/29 & 0.10 & 1.00 & $[-0.20, 2.35]$ \\
Qwen-32B (OR) & base $\to$ + corr. & $+1.37$ & 25/4 & $1.0\times10^{-4}$ & 0.0020 & $[0.52, 2.10]$ \\
Qwen-7B & lexical $\to$ + corr. & $+1.89$ & 35/6 & $4.9\times10^{-6}$ & $1.1\times10^{-4}$ & $[1.06, 2.75]$ \\
Qwen-14B & lexical $\to$ + corr. & $+2.22$ & 35/1 & $1.1\times10^{-9}$ & $2.6\times10^{-8}$ & $[1.50, 3.14]$ \\
Qwen-32B & lexical $\to$ + corr. & $+1.69$ & 32/6 & $2.4\times10^{-5}$ & $4.9\times10^{-4}$ & $[0.46, 3.24]$ \\
Qwen-32B (OR) & embedding $\to$ + corr. & $+1.24$ & 22/3 & $1.6\times10^{-4}$ & 0.0028 & $[0.26, 2.19]$ \\
\midrule
Qwen-7B & base $\to$ lexical & $+0.85$ & 93/80 & 0.36 & 1.00 & $[-2.92, 5.76]$ \\
Qwen-14B & base $\to$ lexical & $-2.15$ & 55/88 & 0.0072 & 0.10 & $[-5.61, 0.42]$ \\
Qwen-32B & base $\to$ lexical & $-1.17$ & 48/66 & 0.11 & 1.00 & $[-3.19, 0.22]$ \\
Llama-8B & base $\to$ lexical & $-2.35$ & 63/99 & 0.0058 & 0.087 & $[-4.83, -0.63]$ \\
Llama-70B & base $\to$ lexical & $-4.24$ & 37/102 & $3.2\times10^{-8}$ & $7.4\times10^{-7}$ & $[-6.98, -1.98]$ \\
Qwen-32B (OR) & base $\to$ lexical & $-1.04$ & 47/63 & 0.15 & 1.00 & $[-3.10, 0.37]$ \\
Qwen-32B (OR) & base $\to$ embedding & $-0.39$ & 33/39 & 0.56 & 1.00 & $[-1.00, 0.25]$ \\
Qwen-32B (OR) & lexical $\to$ embedding & $+0.65$ & 67/57 & 0.42 & 1.00 & $[-1.14, 3.01]$ \\
\midrule
Qwen-32B & + corr. $\to$ + self-consistency & $-0.07$ & 13/14 & 1.00 & 1.00 & $[-0.47, 0.37]$ \\
Qwen-32B & lexical + corr. $\to$ + self-consistency & $+0.13$ & 18/16 & 0.86 & 1.00 & $[-0.93, 1.37]$ \\}
\newcommand{\TabCLBody}{%
CodeLlama-7B & base $\to$ + corr. & $+3.39$ & 55/3 & $2.3\times10^{-13}$ & $6.1\times10^{-12}$ & $[2.36, 4.77]$ \\
CodeLlama-13B & base $\to$ + corr. & $+4.37$ & 69/2 & $2.2\times10^{-18}$ & $6.5\times10^{-17}$ & $[2.65, 6.24]$ \\
CodeLlama-7B & lexical $\to$ + corr. & $+2.74$ & 43/1 & $5.1\times10^{-12}$ & $1.3\times10^{-10}$ & $[1.79, 3.78]$ \\
CodeLlama-13B & lexical $\to$ + corr. & $+3.59$ & 56/1 & $8.0\times10^{-16}$ & $2.3\times10^{-14}$ & $[2.32, 4.92]$ \\
\midrule
CodeLlama-7B & base $\to$ lexical & $+1.69$ & 90/64 & 0.044 & 0.57 & $[-0.42, 3.36]$ \\
CodeLlama-13B & base $\to$ lexical & $+0.07$ & 63/62 & 1.00 & 1.00 & $[-1.93, 2.11]$ \\
CodeLlama-7B & base $\to$ embedding & $-1.11$ & 32/49 & 0.075 & 0.90 & $[-2.02, -0.32]$ \\
CodeLlama-13B & base $\to$ embedding & $-0.59$ & 28/37 & 0.32 & 1.00 & $[-2.41, 0.97]$ \\
CodeLlama-7B & lexical $\to$ embedding & $-2.80$ & 59/102 & $8.8\times10^{-4}$ & 0.014 & $[-4.47, -0.61]$ \\
CodeLlama-13B & lexical $\to$ embedding & $-0.65$ & 64/74 & 0.44 & 1.00 & $[-1.79, 0.87]$ \\}
\newcommand{\TabMechBody}{%
Qwen-7B & 283 & 35 & $+2.28$ $[1.39, 3.36]$ & 33/2 & 34/5 & 326 $\to$ 175 & 426 $\to$ 571 & 1.31 \\
Qwen-14B & 144 & 24 & $+1.56$ $[1.18, 2.06]$ & 24/0 & 2/3 & 147 $\to$ 55 & 420 $\to$ 482 & 1.13 \\
Qwen-32B & 114 & 26 & $+1.69$ $[1.00, 2.31]$ & 25/0 & 2/7 & 116 $\to$ 24 & 421 $\to$ 472 & 1.10 \\
Llama-8B (FP8 API) & 392 & 57 & $+3.72$ $[2.51, 4.99]$ & 52/2 & 19/8 & 393 $\to$ 201 & 402 $\to$ 602 & 1.40 \\
Llama-70B (FP8 API) & 77 & 21 & $+1.37$ $[0.68, 2.14]$ & 17/0 & 27/29 & 80 $\to$ 17 & 400 $\to$ 437 & 1.06 \\
Qwen-32B (OR) & 116 & 25 & $+1.63$ $[0.90, 2.35]$ & 24/0 & 1/4 & 115 $\to$ 24 & 421 $\to$ 476 & 1.10 \\
CodeLlama-7B & 634 & 53 & $+3.46$ $[2.42, 4.82]$ & 53/1 & 2/2 & 632 $\to$ 361 & 557 $\to$ 1012 & 1.70 \\
CodeLlama-13B & 622 & 68 & $+4.43$ $[2.71, 6.30]$ & 68/1 & 1/1 & 621 $\to$ 377 & 504 $\to$ 971 & 1.69 \\
\midrule
Qwen-7B, lexical & 316 & 31 & $+2.02$ $[1.19, 2.90]$ & 31/2 & 4/4 & 316 $\to$ 207 & 357 $\to$ 498 & 1.35 \\
Qwen-14B, lexical & 179 & 33 & $+2.15$ $[1.44, 3.10]$ & 33/0 & 2/1 & 181 $\to$ 73 & 353 $\to$ 425 & 1.17 \\
Qwen-32B, lexical & 128 & 29 & $+1.89$ $[0.81, 3.35]$ & 29/1 & 3/5 & 130 $\to$ 50 & 355 $\to$ 412 & 1.12 \\
Qwen-32B (OR), embedding & 110 & 19 & $+1.24$ $[0.42, 2.11]$ & 19/0 & 3/3 & 111 $\to$ 30 & 393 $\to$ 446 & 1.10 \\
CodeLlama-7B, lexical & 605 & 41 & $+2.67$ $[1.69, 3.77]$ & 41/1 & 2/0 & 603 $\to$ 367 & 463 $\to$ 858 & 1.67 \\
CodeLlama-13B, lexical & 579 & 55 & $+3.59$ $[2.24, 4.98]$ & 55/1 & 1/0 & 583 $\to$ 341 & 416 $\to$ 809 & 1.64 \\}
\newcommand{\TabCostBody}{%
Qwen-7B (self-hosted) & base / + corr. & 426 / 571 & 1.00 / 1.31 & 3.6 / 7.2 & -- \\
Qwen-14B (self-hosted) & base / + corr. & 420 / 482 & 1.00 / 1.13 & 5.9 / 8.7 & -- \\
Qwen-32B (self-hosted) & base / + corr. & 421 / 472 & 1.00 / 1.10 & 7.3 / 10.6 & -- \\
 & lexical / + corr. & 355 / 412 & 1.00 / 1.12 & 7.6 / 13.1 & -- \\
 & + corr. + self-consistency & 2358 & 5.48 & 50.0 & -- \\
CodeLlama-7B (self-hosted) & base / + corr. & 557 / 1012 & 1.00 / 1.70 & -- & -- \\
CodeLlama-13B (self-hosted) & base / + corr. & 504 / 971 & 1.00 / 1.69 & -- & -- \\
\midrule
Llama-8B (FP8 API) & base / + corr. & 402 / 602 & 1.00 / 1.40 & 5.2 / 6.6 & 204 / 293 \\
Llama-70B (FP8 API) & base / + corr. & 400 / 437 & 1.00 / 1.06 & 19.3 / 11.7 & 499 / 348 \\
Qwen-32B (OpenRouter) & base / + corr. & 421 / 476 & 1.00 / 1.10 & 3.6 / 5.2 & 235 / 315 \\
 & embedding / + corr. & 393 / 446 & 1.00 / 1.10 & 3.6 / 5.5 & 409 / 481 \\}
\newcommand{\TabDescBody}{%
Qwen2.5-Coder 7B $\to$ 14B, base & 42.63 & 52.02 & $+9.39$ & 217/73 & $9.3\times10^{-18}$ & $[6.23, 13.68]$ \\
Qwen2.5-Coder 14B $\to$ 32B, base$^{a}$ & 52.02 & 54.89 & $+2.87$ & 133/89 & 0.0038 & $[-0.15, 5.63]$ \\
Llama-3.1-8B $\to$ Llama-3.3-70B, base & 37.09 & 54.69 & $+17.60$ & 352/82 & $6.5\times10^{-41}$ & $[13.22, 21.68]$ \\
CodeLlama 7B $\to$ 13B, base & 23.73 & 26.27 & $+2.54$ & 144/105 & 0.016 & $[-0.56, 5.47]$ \\
\midrule
CodeLlama-7B $\to$ Llama-3.1-8B, base$^{b}$ & 23.73 & 37.09 & $+13.36$ & 300/95 & $7.3\times10^{-26}$ & $[10.14, 16.58]$ \\
CodeLlama-7B $\to$ Qwen2.5-Coder-7B, base$^{a}$ & 23.73 & 42.63 & $+18.90$ & 371/81 & $2.1\times10^{-45}$ & $[14.46, 22.77]$ \\
\midrule
Qwen-32B self-hosted $\to$ OR v4, base & 54.89 & 54.50 & $-0.39$ & 2/8 & 0.11 & $[-0.75, 0.00]$ \\
Qwen-32B self-hosted $\to$ OR v4, + corr. & 56.19 & 55.87 & $-0.33$ & 5/10 & 0.30 & $[-0.74, 0.20]$ \\
Qwen-32B self-hosted $\to$ OR v4, lexical & 53.72 & 53.46 & $-0.26$ & 3/7 & 0.34 & $[-0.59, 0.06]$ \\
Qwen-32B self-hosted $\to$ OR v2 (unrepaired), base & 54.89 & 51.56 & $-3.32$ & 2/53 & $8.6\times10^{-14}$ & $[-4.31, -2.15]$ \\
\midrule
OR v2 $\to$ v3 (replication), base & 51.56 & 51.76 & $+0.20$ & 4/1 & 0.38 & $[-0.12, 0.53]$ \\
OR v2 $\to$ v3 (replication), + corr. & 53.46 & 53.26 & $-0.20$ & 4/7 & 0.55 & $[-0.65, 0.19]$ \\
OR v2 $\to$ v3 (replication), lexical & 50.26 & 50.33 & $+0.07$ & 5/4 & 1.00 & $[-0.14, 0.29]$ \\
OR v2 $\to$ v3 (replication), embedding & 51.11 & 51.24 & $+0.13$ & 4/2 & 0.69 & $[-0.13, 0.39]$ \\
OR v2 $\to$ v3 (replication), emb. + corr. & 52.35 & 52.48 & $+0.13$ & 3/1 & 0.62 & $[-0.13, 0.38]$ \\
\midrule
OR v3 $\to$ v4 (repair), base & 51.76 & 54.50 & $+2.74$ & 44/2 & $3.1\times10^{-11}$ & $[1.71, 3.58]$ \\
OR v3 $\to$ v4 (repair), + corr. & 53.26 & 55.87 & $+2.61$ & 47/7 & $2.3\times10^{-8}$ & $[1.36, 3.61]$ \\
OR v3 $\to$ v4 (repair), lexical & 50.33 & 53.46 & $+3.13$ & 51/3 & $2.9\times10^{-12}$ & $[2.04, 4.10]$ \\
OR v3 $\to$ v4 (repair), embedding & 51.24 & 54.11 & $+2.87$ & 46/2 & $8.4\times10^{-12}$ & $[1.67, 3.81]$ \\
OR v3 $\to$ v4 (repair), emb. + corr. & 52.48 & 55.35 & $+2.87$ & 45/1 & $1.3\times10^{-12}$ & $[1.64, 3.93]$ \\}

\newcommand{\pHQwenFourteenLink}{0.10}

\newcommand{\pHLlamaEightLink}{0.087}
\newcommand{\pHLlamaSeventyCorr}{1.00}
\newcommand{\pHLlamaSeventyLink}{$7.4\times10^{-7}$}

\newcommand{\flipQwenRange}{0.20--2.54}
\newcommand{\flipQwenShort}{0.2--2.5}
\newcommand{\flipORRange}{0.33--0.39}
\newcommand{\halfAllRange}{0.22 to 0.96}
\newcommand{\halfBaseRange}{0.29 to 0.96}
\newcommand{\withinNRange}{19 to 57}
\newcommand{\withinPtsRange}{$+1.24$ to $+3.72$}
\newcommand{\withinNConf}{10}
\newcommand{\withinAllCIpos}{every}
\newcommand{\clSevenBase}{23.73}

\newcommand{\clSevenCorrD}{$+3.39$}

\newcommand{\clSevenCorrH}{$6.1\times10^{-12}$}

\newcommand{\clSevenLinkD}{$+1.69$}
\newcommand{\clSevenLinkP}{0.044}

\newcommand{\clSevenEmbD}{$-1.11$}

\newcommand{\clSevenEmbH}{0.90}

\newcommand{\clSevenEmbCI}{$[-2.02, -0.32]$}
\newcommand{\clSevenEmbLexD}{$-2.80$}

\newcommand{\clSevenEmbLexH}{0.014}

\newcommand{\clSevenWithin}{$+3.46$}
\newcommand{\clSevenWithinCI}{$[2.42, 4.82]$}

\newcommand{\clSevenTokBase}{557}

\newcommand{\clSevenTokCorr}{1012}

\newcommand{\clSevenTokRatio}{1.82}
\newcommand{\clSevenEM}{2.28}
\newcommand{\clThirteenBase}{26.27}

\newcommand{\clThirteenCorrD}{$+4.37$}

\newcommand{\clThirteenCorrH}{$6.5\times10^{-17}$}

\newcommand{\clThirteenLinkD}{$+0.07$}

\newcommand{\clThirteenEmbD}{$-0.59$}

\newcommand{\clThirteenEmbH}{1.00}

\newcommand{\clThirteenEmbLexD}{$-0.65$}

\newcommand{\clThirteenEmbLexH}{1.00}

\newcommand{\clThirteenWithin}{$+4.43$}
\newcommand{\clThirteenWithinCI}{$[2.71, 6.30]$}

\newcommand{\clThirteenTokBase}{504}

\newcommand{\clThirteenTokCorr}{971}

\newcommand{\clThirteenTokRatio}{1.93}
\newcommand{\clThirteenEM}{2.41}
\newcommand{\clGapLlama}{13.36}

\newcommand{\clGapQwen}{18.90}

\newcommand{\clSizeD}{$+2.54$}
\newcommand{\clSizeP}{0.016}
\newcommand{\clSizeCI}{$[-0.56, 5.47]$}
\newcommand{\clTimedOutTotal}{832}
\newcommand{\clTimedOutMin}{7}
\newcommand{\clTimedOutMax}{331}
\newcommand{\clTimedOutMaxRun}{\texttt{codellama-13b-clean.correct}}
\newcommand{\clStillFailed}{1}
\newcommand{\clRetryConc}{32}
\newcommand{\clRetryTimeout}{900}
\newcommand{\clVllm}{0.31.0}
\newcommand{\clFlipRange}{0.07--0.26}
\newcommand{\clFlipRangeShort}{0.07--0.26}
\newcommand{\clHalfRange}{0.13 to 0.26}
\newcommand{\clWithinRange}{41 to 68 items, $+2.67$ to $+4.43$ points}

\title{How far do on-premises open LLMs get on Text-to-SQL? A cross-family size and technique frontier on BIRD}

\author[1]{Vladimir Beskorovainyi\,\orcidlink{0009-0004-7005-6242}}
\affil[1]{Besk Tech, Moscow, Russia}

\corrauthor[1]{Vladimir Beskorovainyi}{admin@besk.tech}

\begin{abstract}
Organizations that cannot send data to a cloud API need to know how well open-weight language models perform on Text-to-SQL when they serve the models themselves, and which inexpensive accuracy techniques justify their compute. We report a controlled benchmark on the BIRD development split ($n=1534$) with a single zero-shot prompt, scored with the official BIRD execution-accuracy comparison on read-only databases. We evaluate Qwen2.5-Coder-Instruct (7B, 14B, 32B) and CodeLlama-Instruct (7B, 13B) self-hosted at 16-bit precision, Llama-3.1-8B and Llama-3.3-70B through an FP8 API, and Qwen2.5-Coder-32B through a second API. We ablate lexical and embedding-based schema linking, execution-guided self-correction and self-consistency, and test every within-model recipe difference with paired exact McNemar tests, Holm correction over \holmM{} tests and a database-clustered bootstrap. This article corrects harness defects present in the preprint and an output defect of the second API provider; all reported numbers are re-scored or come from re-runs. Qwen2.5-Coder reaches 42.6\%, 52.0\% and 54.9\% execution accuracy at 7B, 14B and 32B; the step from 14B to 32B ($+2.9$ points) coincides with a change of hardware and serving software, and its clustered confidence interval includes zero. CodeLlama-7B and -13B reach \clSevenBase\% and \clThirteenBase\%. Self-correction was the only component with a consistent gain. Measured within the self-correction run, which removes run-to-run noise, repairing non-executing drafts added 1.2 to 3.7 points for the Qwen2.5-Coder and Llama models (\clSevenWithin{} and \clThirteenWithin{} points for CodeLlama-7B and -13B) at 1.1 to 1.5 times the tokens (\clSevenTokRatio{} and \clThirteenTokRatio{} times for CodeLlama, whose drafts fail more often); the base versus self-correction comparison was significant after Holm correction for every model except Llama-3.3-70B. Schema linking never improved accuracy significantly: lexical linking reduced it for Llama-3.3-70B ($-4.2$ points), and an embedding linker with higher gold-table recall (96.5\% versus 94.0\%) did not differ detectably from no linking ($-0.4$ points, $p=0.56$). Self-consistency with five samples, run for one model, changed accuracy by $-0.07$ and $+0.13$ points at about five times the tokens. The second API provider dropped spaces next to digits; after a whitespace repair, Qwen2.5-Coder-32B scored 54.5\% through that API and 54.9\% self-hosted ($p=0.11$), so the apparent serving gap of about three points in the unrepaired runs was mostly this defect, and two API runs with the same configuration agreed within 0.2 points. A second greedy pass changed the correctness of at least \flipQwenShort\% of items for Qwen2.5-Coder self-hosted (\clFlipRangeShort\% for CodeLlama) and 1.8--3.7\% through the FP8 API, so differences below about one point should not be interpreted. We release the code, the per-item predictions and the analysis scripts.
\end{abstract}

\begin{document}

\flushbottom
\maketitle
\thispagestyle{empty}

\section*{Introduction}

Text-to-SQL is among the most directly useful enterprise applications of large language models (LLMs), yet the strongest published results rely on frontier cloud APIs or on extensive fine-tuning \citep{hongsurvey,liusurvey}. Many organizations cannot send their data or schemas to an external API for reasons of data residency, privacy or cost. For them, the practical question is how well open-weight models perform when they serve the models themselves, and which of the widely used accuracy techniques are worth the additional compute.

Throughout this article, \emph{on-premises} (as in the title) is used in the sense of \emph{self-hosted}: the operator runs the open model weights with an inference server it controls, rather than calling a model provider's API. Our self-hosted runs used rented cloud GPUs and free Kaggle notebook GPUs as stand-ins for an operator's own hardware, not hardware in our own data centre; the accuracy of a model under a given serving stack does not depend on who owns the GPU, but the data-control argument above applies only to hardware under the operator's control.

We address this question empirically on the BIRD development split \citep{bird}. We evaluate three open model families, Qwen2.5-Coder \citep{qwen25coder}, CodeLlama \citep{codellama} and Llama-3.x \citep{llama3}, self-hosted with vLLM \citep{vllm} or, where stated, through commercial APIs. On top of a single zero-shot call, we ablate a model-agnostic recipe one component at a time: schema linking (a lexical pruner and an embedding retriever), execution-guided self-correction and self-consistency \citep{selfconsistency}. Every within-model difference between recipe configurations is tested with the paired exact McNemar test \citep{mcnemar} with Holm correction \citep{holm1979}, accompanied by a bootstrap that accounts for the clustering of questions within databases, and negative results are reported in full.

The study addresses four research questions:
\begin{description}[leftmargin=2.2em,itemsep=2pt]
  \item[RQ1] How does execution accuracy change with model size within a family, and how large are the differences between families? (descriptive; see Methods for the confounds)
  \item[RQ2] Does execution-guided self-correction improve accuracy, and through which mechanism?
  \item[RQ3] Does schema linking improve accuracy, and does a higher-recall embedding linker change the answer?
  \item[RQ4] What do the recipe components cost in tokens, model calls and, where meaningful, wall-clock time?
\end{description}

The contributions are as follows.
\begin{enumerate}[leftmargin=1.4em,itemsep=2pt]
  \item A controlled benchmark of open model families on BIRD under one prompt and one evaluation protocol, with the serving configuration of every run documented as far as it was logged (Table~\ref{tab:config} states what was not logged), and every within-model recipe comparison tested with paired significance tests.
  \item A component-level ablation showing that self-correction is the only component with a consistent gain, that the gain comes from repairing non-executing queries, and that schema linking and self-consistency (the latter tested on one model) yield no measurable gain on BIRD.
  \item An analysis of serving effects: an output defect of one API provider and its repair, a cross-serving comparison of the same model after the repair, a replication of an API run, and a lower-bound estimate of run-to-run nondeterminism, which together bound how small an accuracy difference can be interpreted.
  \item A release of the harness, the per-item predictions and the analysis scripts; every reported number can be recomputed from the released predictions, whereas regeneration from scratch is subject to serving nondeterminism.
\end{enumerate}

\section*{Related work}

\paragraph{Benchmarks.} Spider \citep{spider} established cross-database Text-to-SQL. BIRD \citep{bird} added larger and noisier databases, external-knowledge evidence and an execution-based metric. Spider 2.0 \citep{spider2} moves towards enterprise workflows with very wide schemas and several SQL dialects, and BIRD-CRITIC \citep{birdcritic} targets the diagnosis of user SQL issues. Recent audits report substantial annotation errors in BIRD \citep{birderrors}, which limits the precision with which absolute scores can be interpreted.

\paragraph{Prompting and pipelines.} Early studies evaluated zero-shot and few-shot prompting of LLMs for Text-to-SQL \citep{rajkumar,chang2023prompt}, and DAIL-SQL \citep{dailsql} systematized prompt engineering. Pipelines combine schema linking \citep{resdsql}, decomposition \citep{dinsql}, self-debugging \citep{selfdebug}, self-consistency \citep{selfconsistency} and chain-of-thought prompting \citep{cot}. Multi-agent and multi-candidate systems such as MAC-SQL \citep{macsql}, CHESS \citep{chess}, CHASE-SQL \citep{chasesql} and XiYan-SQL \citep{xiyansqlpreview,xiyansql} add candidate generation and selection on top of these components. Open models fine-tuned for the task, such as CodeS \citep{codes} and OmniSQL \citep{omnisql}, and reinforcement-learning approaches such as SQL-R1 \citep{sqlr1}, Reasoning-SQL \citep{reasoningsql} and Arctic-Text2SQL-R1 \citep{arctic}, reach considerably higher BIRD scores than the zero-shot setting studied here.

\paragraph{Schema linking.} The value of schema linking is contested. RSL-SQL \citep{rslsql} reports gains from bidirectional linking combined with safeguards against recall loss, whereas \citet{deathlinking} and \citet{longcontext} argue that strong models with long context windows can work directly from the full schema. Our study adds a controlled measurement for small and mid-sized open models, with a lexical and an embedding linker, under paired testing.

\paragraph{Open models and serving.} Code-specialized open models \citep{qwen25coder,codellama} and efficient serving \citep{vllm} make private Text-to-SQL feasible. Quantization can change code-generation accuracy \citep{quantcode}, and greedy decoding under batched inference is not bit-reproducible because floating-point reductions depend on batch composition \citep{he2025nondeterminism}. We therefore keep self-hosted 16-bit results and API results in separate blocks and compare across them only where explicitly labelled as cross-serving.

\section*{Materials and methods}

\paragraph{Dataset and preprocessing.} We evaluate on the full BIRD development split ($n=1534$ question--SQL pairs over 11 SQLite databases), using the \texttt{dev\_20240627} package of the official development data; its gold SQL is identical to that used by the original runs for all 1534 items. The BIRD authors have since released a revised development set \citep{birdlb}; we did not use it, so our numbers are comparable to results on the \texttt{dev\_20240627} package. No preprocessing is applied to questions or gold queries. Each database schema is serialized into the prompt as a listing of every table with its column names, followed by the question and the external-knowledge evidence supplied by BIRD. The serialization is deliberately minimal: column names that contain spaces or parentheses are listed unquoted, and no column types, keys or sample values are given (see Limitations). The serializer and prompt templates are released with the code. The databases are obtained from the BIRD authors and are not redistributed.

\paragraph{Evaluation metric.} Execution accuracy (EX) counts a prediction as correct if and only if the set of rows it returns equals the set of rows returned by the gold query, as in the official BIRD evaluation script (\texttt{set(pred) == set(gold)})\footnote{\url{https://github.com/AlibabaResearch/DAMO-ConvAI/tree/main/bird}}. Queries are executed on read-only connections (\texttt{file:\dots?mode=ro}) to pristine copies of the databases, with a 30\,s timeout. A prediction that fails to execute or times out is wrong. A scoring pass is the execution of the gold and predicted queries from which the reported correctness of a run is taken (the official re-scoring for the released runs and the harness evaluation for the re-runs). Because gold results were computed in every scoring pass rather than cached offline, a few slow gold queries exceeded the time limit in some passes but not in others. To keep paired comparisons free of such load-dependent artefacts, we apply one rule to every run: the \goldFailN{} questions whose gold query failed or timed out in at least one scoring pass of any run reported here (question ids \goldFailIDs; 518 and 701 failed in every scoring pass in which their gold query was executed, the others only in some passes, and each pass had 1--5 such items) are scored as wrong in every run. None of these questions was scored correct in any run, so the rule changes no reported EX value, but it guarantees that they never enter a discordant pair. Every configuration is scored on all 1534 items.

\paragraph{Models and serving.} Table~\ref{tab:config} lists every serving configuration. Qwen2.5-Coder-Instruct (7B, 14B, 32B) was self-hosted with vLLM in the checkpoint's 16-bit dtype (vLLM \texttt{dtype=auto}; the dtype actually used was not logged, and the Qwen2.5 checkpoints are distributed in bf16). The 7B and 14B models ran on an L40S GPU with vLLM 0.8.5, and the 32B model on an A100 with vLLM 0.11.1, so model size and serving stack change together between 14B and 32B. CodeLlama-Instruct (7B, 13B) was re-run in fp16 on two Tesla T4 GPUs with tensor parallelism in Kaggle notebook sessions; the T4 does not support bf16, so the checkpoints were cast to fp16. Llama-3.1-8B-Instruct and Llama-3.3-70B-Instruct were served by a commercial API at FP8 precision (DeepInfra). Qwen2.5-Coder-32B-Instruct was additionally run through OpenRouter. At the time of the runs (October 2026) OpenRouter offered a single provider for this model, Cloudflare, which served every call of every stage; the serving precision is not disclosed. We report three blocks, self-hosted 16-bit, FP8 API and OpenRouter API, and draw conclusions within blocks; any comparison across blocks is labelled cross-serving. Decoding is greedy (temperature 0) except for self-consistency sampling.

\begin{table}[ht]
\centering
\caption{Run configuration. Concurrency is the number of questions processed in parallel by the harness. Common to all runs: one zero-shot prompt, greedy decoding, 30\,s execution timeout, official set comparison on read-only databases. Not logged: the dtype selected by vLLM for the Qwen2.5-Coder runs and wall-clock time of the released Qwen2.5-Coder runs.}
\label{tab:config}
\footnotesize
\setlength{\tabcolsep}{3pt}
\begin{tabular}{p{2.0cm}p{1.75cm}p{4.15cm}p{1.45cm}ccp{1.3cm}}
\toprule
Models & Serving & Hardware / software & Precision & Conc. & Max tokens & Wall-clock \\
\midrule
Qwen2.5-Coder 7B, 14B & self-hosted vLLM & 1$\times$L40S 48\,GB; vLLM 0.8.5, torch 2.6 & 16-bit (checkpoint dtype) & 64 & 1024 & not logged \\
Qwen2.5-Coder 32B & self-hosted vLLM & 1$\times$A100-SXM4 80\,GB; vLLM 0.11.1, torch 2.9 & 16-bit (checkpoint dtype) & 64 & 1024 & not logged \\
CodeLlama 7B, 13B (re-run) & self-hosted vLLM (Kaggle) & 2$\times$Tesla T4 16\,GB, tensor parallel 2; vLLM \clVllm{}, eager mode, prefix caching, \texttt{max\_model\_len} 3072, \texttt{max\_num\_seqs} 64; client timeout 90\,s with 2 retries; retry pass at concurrency \clRetryConc{}, timeout \clRetryTimeout\,s & fp16 & 64 & 1024 & not used$^{a}$ \\
Llama-3.1-8B, Llama-3.3-70B & DeepInfra API & provider-managed & FP8 & 24 & 1024 & logged \\
Qwen2.5-Coder 32B (re-run) & OpenRouter API (Cloudflare) & provider-managed; whitespace repair of every completion (see text) & undisclosed & 16 & 4096$^{b}$ & logged \\
\bottomrule
\end{tabular}

\smallskip
{\footnotesize $^{a}$ The first-pass wall clock includes requests that timed out on the client and excludes the retry pass. $^{b}$ No self-hosted Qwen2.5-Coder-32B base completion exceeded 437 tokens, so the larger limit does not affect the comparison.}
\end{table}

\paragraph{API output repair.} The OpenRouter completions contain a provider-side defect: spaces adjacent to digits are dropped, so that, for example, \texttt{> 500 AND} arrives as \texttt{>500AND} and \texttt{LIMIT 1} as \texttt{LIMIT1}. In two unrepaired runs of all five stages (described below), 81 to 88 final predictions per stage contained a digit immediately followed by an upper-case SQL keyword; none of them executed, and the self-hosted runs of the same model contained no such prediction. For the main OpenRouter arm we therefore apply a whitespace repair to every completion before SQL extraction: outside quoted literals and identifiers, a space is inserted between a digit and an adjacent upper-case SQL keyword (\texttt{src/nl2sql\_bench/textfix.py}). The repair only inserts whitespace at token boundaries that SQLite cannot parse, and it modified 97 to 120 completions per stage. With the repair, no final prediction contains a digit glued to a keyword, and no OpenRouter call failed (0 runtime errors in all five stages). We found the defect in our first OpenRouter run of this study (labelled v2 in the released files; v1 denotes the preprint's API arm), which had no repair, and in a replication with identical configuration (v3; its repair step was inadvertently inactive). The reported OpenRouter arm (v4) applies the repair; v2 and v3 are released and are used only to quantify the effect of the repair and as a replication of an API run.

\paragraph{Recipe and ablation.} On top of single-call generation we ablate three components.
\emph{Lexical schema linking} keeps a whole table (all of its columns) if any word-level sub-token of its name or of one of its column names occurs in the question or the evidence; it performs no primary/foreign-key closure and no column pruning, and it keeps the full schema if nothing matches.
\emph{Embedding schema linking} embeds each table description (table name and column names) with the all-MiniLM-L6-v2 sentence encoder \citep{sbert,minilm} and keeps the top $k=6$ tables by cosine similarity to the question and evidence ($k$ is recorded in the run summaries; the code default is 5). Because five of the eleven development databases have at most six tables, it keeps the full schema for 652 of 1534 questions (42.5\%). The value $k=6$ was chosen on the evaluation split (see Limitations).
\emph{Self-correction} executes the draft; if execution fails or times out, the model receives the error message and is asked to repair the query, for up to two repair rounds. The repair prompt omits the external-knowledge evidence. In the released Qwen2.5-Coder self-hosted runs, produced with an earlier harness revision, repair was triggered by execution errors only, not by timeouts.
\emph{Self-consistency} draws five samples at temperature 0.6; each sample is self-corrected and executed, candidates are grouped by execution result, the largest group wins, ties are broken by the earliest candidate, and the greedy candidate is not included. The grouping key is the sorted multiset of row representations, which is slightly stricter than the set comparison of the metric. Only the selected query was stored, so pass@5 and oracle-selection accuracy cannot be computed. Self-consistency was run only for Qwen2.5-Coder-32B self-hosted.

\paragraph{Significance testing.} All configurations are scored on the same items, so differences are paired. We use the exact McNemar test \citep{mcnemar} on per-item correctness (a two-sided binomial test on the discordant pairs) and control the family-wise error rate with the Holm procedure \citep{holm1979} over all \holmM{} within-model recipe comparisons reported (\holmMain{} for the Qwen2.5-Coder, Llama and OpenRouter runs and \holmCL{} for CodeLlama). McNemar's test treats items as independent, whereas BIRD questions are nested in 11 databases. As a sensitivity analysis we therefore report, for every paired difference, a database-clustered percentile bootstrap 95\% confidence interval (the 11 databases are resampled with replacement, 10,000 replicates, seed 0). Comparisons between model sizes, between families, across serving stacks and between replicate runs are reported descriptively with the same two statistics; they are not part of the Holm family and are not used for confirmatory claims.

\paragraph{Self-correction estimands and serving nondeterminism.} A self-correction run issues the base prompt first and repairs only drafts that fail to execute. Comparing the base run with the self-correction run therefore mixes two effects: the repairs, and any difference between two greedy generations of the same prompt in separate runs. Our primary estimand for self-correction is the within-run effect: the number of items whose draft failed to execute (and was therefore wrong) and whose repaired final query is correct, divided by 1534. It does not involve a second run and is free of run-to-run variation; by construction it cannot be negative, so we report it with a clustered bootstrap interval rather than a test. The base versus self-correction McNemar test is reported as a secondary analysis. On items where no repair was triggered, the self-correction run reissued the base prompt verbatim, so these items provide a second greedy sample of the base configuration. The fraction of all 1534 items whose correctness changes on these items is a lower bound on run-to-run variation, because unrepaired items are the easier ones, and it mixes symmetric serving noise with any systematic difference between the two runs. With $k$ symmetric flips, the standard deviation of a difference in EX between two runs is about $\sqrt{k}/1534$.

\paragraph{Contamination probe.} As a cheap probe for verbatim memorization we compute, for every run, the share of predictions that equal the gold query after lower-casing, collapsing whitespace and removing a trailing semicolon.

\paragraph{Cost.} We report mean tokens per question (prompt plus completion over all model calls) and mean model calls for every run. Per-item latency is the summed duration of the model calls for that item under concurrent load, including client-side retries; its 95th percentile (p95) describes the load on the serving stack, not the latency experienced by a single user. Wall-clock time is reported where it was logged and is meaningful. We do not convert any measurement into dollars: the released self-hosted Qwen2.5-Coder runs did not log wall-clock time, the CodeLlama first-pass wall clock is contaminated by client timeouts, and API timings reflect a shared, provider-managed service.

\paragraph{Corrections relative to the preprint.} The first version of the preprint (arXiv:2606.29733v1) was affected by six defects, all of which are corrected here. (1) The harness opened the BIRD databases read-write, so data-definition statements emitted by models (for example \texttt{ALTER TABLE \dots{} ADD COLUMN}) executed and persisted; 68 final CodeLlama predictions contained such statements, and the CodeLlama-13B and -34B runs saw altered schemas. All scoring and re-runs now use read-only connections. (2) The evaluator compared sorted multisets of row representations, which is stricter than the official BIRD script; all EX values are now computed with the official set comparison on pristine databases, which raises the EX of the 20 retained released runs by 3.6 to 5.5 points. (3) The FP8-API arm of the preprint for Qwen2.5-Coder-32B was invalid: 508 of 1534 completions were truncated at 1024 tokens of chain-of-thought prose. It is replaced by an OpenRouter re-run (maximum 4096 tokens, no truncated completions), and the preprint's claim of a quantization penalty of about 13 points is withdrawn. (4) The SQL extractor could start inside prose; it now takes the last fenced block, else the first line starting with \texttt{SELECT} or \texttt{WITH}, else the first \texttt{SELECT} or common-table-expression header. By a heuristic count, the old extractor affected 0 to 5 items per released run; the retained released runs are re-scored as produced, and all re-runs use the new extractor. The new extractor can still start inside prose when a prose line begins with ``With'' or ``Select'', and it takes the last fenced block even if that block is not the answer; we found no material effect on the re-runs. (5) The CodeLlama-34B embedding-linker run of the preprint (\texttt{codellama-34b.link\_emb}) lost 61 items to a race when loading the embedder; the embedder is now loaded once before the worker threads start. (6) Items with a failing gold query are now counted as wrong by one uniform rule (Evaluation metric). As a consequence, the CodeLlama results of the preprint are replaced by clean re-runs of the 7B and 13B models, CodeLlama-34B is no longer reported, and two significance claims of the preprint no longer hold: lexical linking does not significantly reduce accuracy for Qwen2.5-Coder-32B or Llama-3.1-8B after correction.

\paragraph{CodeLlama retry pass.} In the first pass of the CodeLlama re-runs, 64 concurrent requests on two T4 GPUs with a 90\,s client timeout caused \clTimedOutMin{} to \clTimedOutMax{} requests per run (\clTimedOutTotal{} in total; most in \clTimedOutMaxRun) to time out on the client; these items were re-issued with the same prompts, greedy decoding and harness at concurrency \clRetryConc{} with a \clRetryTimeout\,s timeout, and each retried record replaced the first-pass record. After the retry pass, \clStillFailed{} item in the CodeLlama-13B self-correction run still failed: its draft reached the 1024-token completion limit, and the resulting repair prompt (2049 input tokens plus 1024 requested output tokens) exceeded the 3072-token context limit of the server, which rejected the request in all three retry attempts. It is scored wrong; all other retried items completed. The OpenRouter runs had no failed requests.

\section*{Results}

Table~\ref{tab:ex} gives EX for every configuration, and Table~\ref{tab:sig} the paired tests.

\begin{table}[ht]
\centering
\caption{Execution accuracy (\%) on the BIRD development split ($n=1534$), official set comparison on read-only databases. Blocks differ in serving and are not directly comparable. ``Corr.'': self-correction; ``--'': not run. The OpenRouter row is the whitespace-repaired run. Self-consistency on Qwen2.5-Coder-32B (self-hosted): 56.13 after self-correction and 55.54 after lexical linking and self-correction.}
\label{tab:ex}
\small
\setlength{\tabcolsep}{4pt}
\begin{tabular}{llcccccc}
\toprule
 & & & & Lexical & Lexical & Embedding & Embedding \\
Serving & Model & Base & + Corr. & linking & + Corr. & linking & + Corr. \\
\midrule
\TabEXBody
\bottomrule
\end{tabular}
\end{table}

\begin{table}[ht]
\centering
\caption{Paired exact McNemar tests for the Qwen2.5-Coder, Llama and OpenRouter runs. $\Delta$: difference in EX (percentage points); $+/-$: items gained/lost; $p_{\mathrm{Holm}}$: Holm-adjusted over all \holmM{} recipe comparisons (including the \holmCL{} CodeLlama comparisons in Table~\ref{tab:cl}); CI: database-clustered bootstrap 95\% interval for $\Delta$. OR: OpenRouter, whitespace-repaired.}
\label{tab:sig}
\footnotesize
\setlength{\tabcolsep}{4pt}
\begin{tabular}{llrrrrr}
\toprule
Model & Comparison & $\Delta$ & $+/-$ & $p$ & $p_{\mathrm{Holm}}$ & Clustered 95\% CI \\
\midrule
\TabSigBody
\bottomrule
\end{tabular}
\end{table}

\subsection*{RQ1: size and family}

\begin{figure}[ht]
\centering
\includegraphics[width=0.85\linewidth]{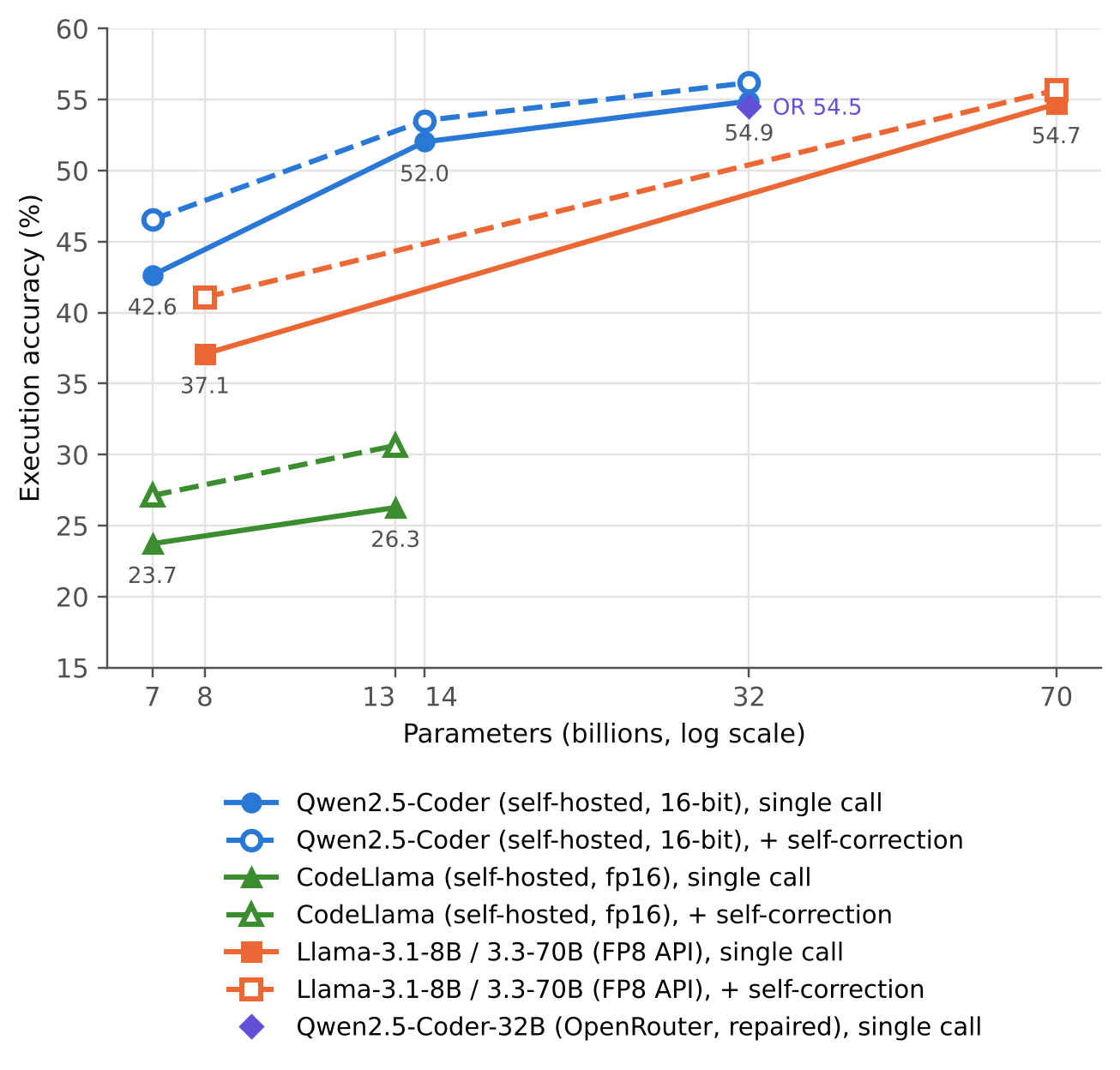}
\caption{Execution accuracy against model size (log scale) on the BIRD development split. Blue circles: Qwen2.5-Coder, self-hosted at 16-bit (7B and 14B on L40S with vLLM 0.8.5, 32B on A100 with vLLM 0.11.1); green triangles: CodeLlama, self-hosted in fp16 on two T4 GPUs; orange squares: Llama-3.1-8B and Llama-3.3-70B (different releases) through an FP8 API; violet diamond: Qwen2.5-Coder-32B through OpenRouter, whitespace-repaired. Filled markers and solid lines: single call; open markers and dashed lines: with self-correction. Labels give base EX. Lines connect sizes within a family and do not imply interpolation; series from different serving blocks are cross-serving with respect to each other.}
\label{fig:size}
\end{figure}

Size and family contrasts are descriptive (Table~\ref{tab:desc}). Within the self-hosted block, Qwen2.5-Coder base EX rises from 42.63\% (7B) to 52.02\% (14B) and 54.89\% (32B), that is, by 9.39 points and then by 2.87 points for a further doubling of parameters (Figure~\ref{fig:size}). The first step was measured on one serving stack and its clustered interval is far from zero. The second step coincides with a change of GPU and vLLM version (L40S with vLLM 0.8.5 for 14B, A100 with vLLM 0.11.1 for 32B), its unadjusted McNemar $p$ is 0.0038 and its clustered interval ($-0.15$ to $5.63$ points) includes zero. The pattern is consistent with diminishing returns to size, but the 14B to 32B increment is not separately attributable to model size. Within the FP8 API block, Llama-3.3-70B (54.69\%) is 17.60 points above Llama-3.1-8B (37.09\%); the two models belong to different releases. CodeLlama-Instruct reaches \clSevenBase\% at 7B and \clThirteenBase\% at 13B on one serving stack; the difference of \clSizeD{} points has an unadjusted $p$ of \clSizeP{}, but its clustered interval (\clSizeCI) includes zero.

\paragraph{Generation and family: a suggestive, untested contrast.} CodeLlama is derived from Llama~2 (2023), whereas Qwen2.5-Coder and Llama-3.x are 2024 releases. The closest within-lineage pair is Llama-3.1-8B and CodeLlama-7B; their base EX differs by \clGapLlama{} points. This contrast is cross-serving (FP8 API versus fp16 on T4 GPUs) and confounds generation with code specialization, training data and possible exposure to BIRD during training (see the contamination probe below), so it is suggestive rather than a test of a generation effect. The contrast between Qwen2.5-Coder-7B and CodeLlama-7B (\clGapQwen{} points) compares two families and two generations at once, on different GPUs and vLLM versions, and cannot separate these factors either.

\paragraph{Serving: the same model through an API.} After the whitespace repair, Qwen2.5-Coder-32B scored 54.50\% through OpenRouter and 54.89\% self-hosted in the base configuration ($-0.39$ points; 2 items gained and 8 lost; exact McNemar $p=0.11$; clustered CI $-0.75$ to $0.00$), and the differences after self-correction ($-0.33$ points, $p=0.30$) and after lexical linking ($-0.26$ points, $p=0.34$) were of the same size. The unrepaired run v2 showed a gap of 3.32 points (2 gained, 53 lost). Of the 53 lost items, 45 differed from the self-hosted prediction only in whitespace and 43 of these did not execute; across all base predictions, 224 of the 288 that differed from the self-hosted run differed only in whitespace. The repair raised OpenRouter EX by 2.61 to 3.13 points per stage relative to the identical unrepaired run (Table~\ref{tab:desc}). Most of the apparent serving gap of the unrepaired runs was therefore an output defect of one provider rather than an effect of precision or inference kernels; what remains after the repair is a cross-serving difference of less than half a point whose interval includes zero, and it is of the order of run-to-run noise. The two unrepaired OpenRouter runs replicate each other closely: 1459 to 1473 of 1534 final predictions per stage were identical, and the five stage-wise EX differences ranged from $-0.20$ to $+0.20$ points ($p\ge0.38$).

\begin{table}[ht]
\centering
\caption{Descriptive paired comparisons, not included in the Holm family. $+/-$: items gained/lost from the first to the second configuration; $p$: unadjusted exact McNemar; CI: database-clustered bootstrap 95\% interval. $^{a}$Different GPU and vLLM version. $^{b}$Different serving block (FP8 API versus self-hosted fp16). OR v2/v3: unrepaired OpenRouter runs with identical configuration; OR v4: whitespace-repaired run used in the main tables.}
\label{tab:desc}
\footnotesize
\setlength{\tabcolsep}{2.8pt}
\begin{tabular}{lrrrrrr}
\toprule
Comparison & EX$_1$ & EX$_2$ & $\Delta$ & $+/-$ & $p$ & Clustered 95\% CI \\
\midrule
\TabDescBody
\bottomrule
\end{tabular}
\end{table}

\subsection*{RQ2: self-correction}

Within the self-correction runs, repairs of non-executing drafts turned \withinNRange{} items into correct answers for the Qwen2.5-Coder, Llama and OpenRouter runs, that is, \withinPtsRange{} points over all \withinNConf{} of their self-correction configurations, and \withinAllCIpos{} clustered interval in Table~\ref{tab:mech}, including the CodeLlama rows, excludes zero (Table~\ref{tab:mech}; for example $+1.69$ points, CI $1.00$ to $2.31$, for Qwen2.5-Coder-32B and $+1.37$ points, CI $0.68$ to $2.14$, for Llama-3.3-70B). For CodeLlama the within-run gains are \clSevenWithin{} (7B, CI \clSevenWithinCI) and \clThirteenWithin{} points (13B, CI \clThirteenWithinCI) from the base configuration, and \clWithinRange{} over all four CodeLlama self-correction configurations. Self-correction sharply reduces the number of non-executing final queries (for example from 116 to 24 for Qwen2.5-Coder-32B).

\begin{figure}[p]
\centering
\includegraphics[width=0.95\linewidth]{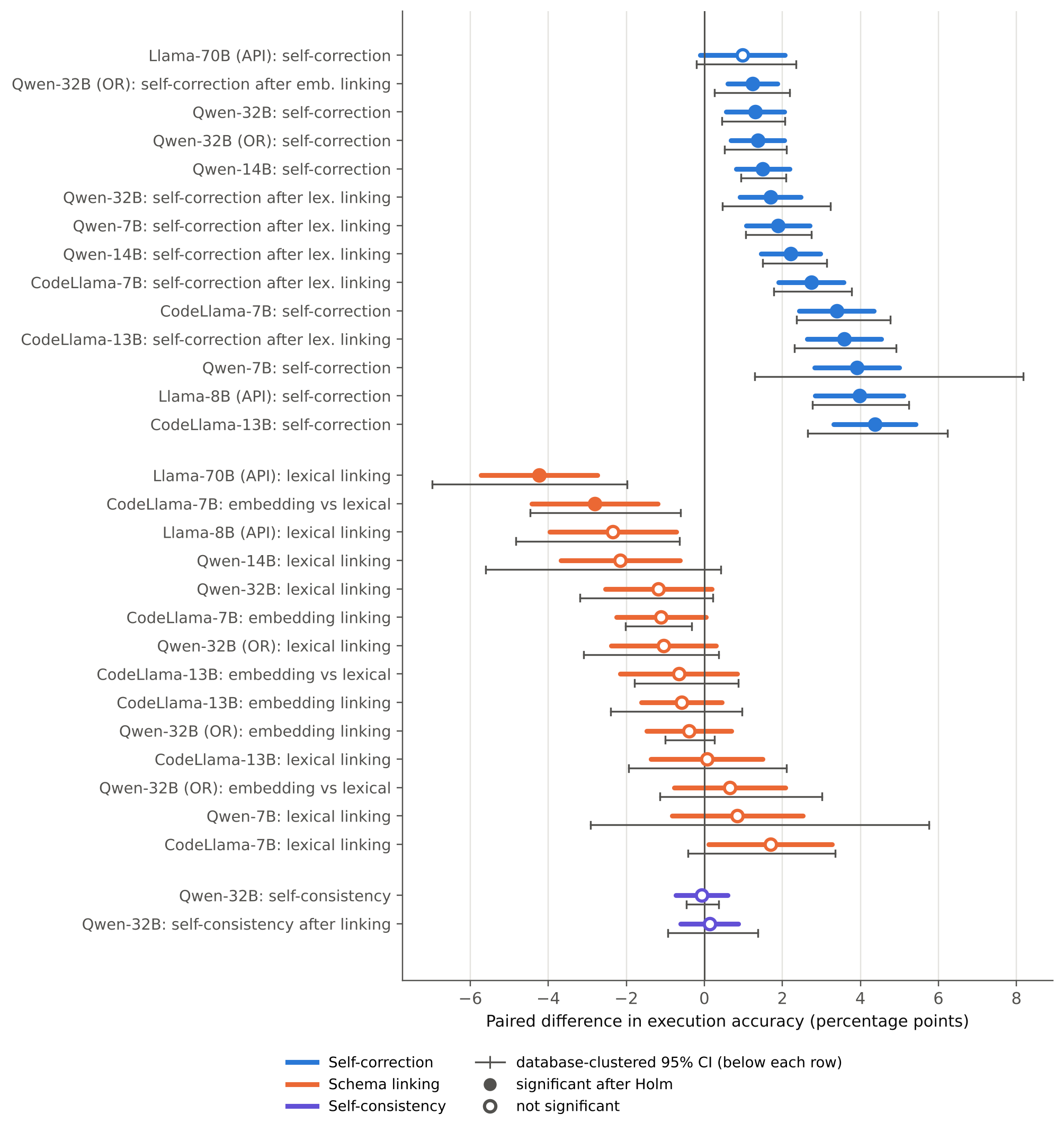}
\caption{Paired differences in execution accuracy for every comparison in the Holm family, grouped by recipe component. Thick bars: unadjusted 95\% confidence intervals for a paired difference in proportions; thin grey bars with end caps, drawn just below each row: database-clustered bootstrap 95\% intervals; filled markers indicate significance after Holm correction over the \holmM{} comparisons, open markers non-significance. ``(API)'': FP8 API serving; ``(OR)'': OpenRouter, whitespace-repaired; all other rows are self-hosted at 16-bit.}
\label{fig:effects}
\end{figure}

In the secondary comparison of the base run with the self-correction run, self-correction improved EX in every configuration in which it was run, and the improvement is significant after Holm correction for Qwen2.5-Coder at all three sizes, for Llama-3.1-8B and for the OpenRouter run, both from the base and from the linked configurations (Table~\ref{tab:sig}, Figure~\ref{fig:effects}), with gains from $+1.24$ to $+3.98$ points. For Llama-3.3-70B the difference ($+0.98$ points, $p=0.10$, $p_{\mathrm{Holm}}=$\,\pHLlamaSeventyCorr) is not significant. For CodeLlama the differences are \clSevenCorrD{} points at 7B ($p_{\mathrm{Holm}}=$\,\clSevenCorrH) and \clThirteenCorrD{} points at 13B ($p_{\mathrm{Holm}}=$\,\clThirteenCorrH), and the lexical-linking pairs are also significant (Table~\ref{tab:cl}); the base versus self-correction comparison is therefore significant after Holm correction for every model except Llama-3.3-70B.

The two estimands differ because the base versus self-correction comparison also contains run-to-run variation on the items that were not repaired. On these items the self-correction run reissued the base prompt, and the final SQL string differed from the base run for 4.0\% (Qwen2.5-Coder-32B), 5.0\% (14B), 10.5\% (7B), 14.3\% (Llama-3.1-8B), 38.9\% (Llama-3.3-70B) and 3.5\% (OpenRouter) of them. Correctness changed for at least \flipQwenRange\% of all 1534 items for Qwen2.5-Coder self-hosted, \flipORRange\% for the OpenRouter run, 1.76\% for Llama-3.1-8B and 3.65\% for Llama-3.3-70B on the FP8 API (\clFlipRange\% for CodeLlama). For Llama-3.3-70B these flips (27 items gained and 29 lost without repair) mask the repair effect in the overall test, although the within-run effect is clearly positive. For Qwen2.5-Coder-7B the unrepaired items moved by $+34/-5$ between the two runs (two-sided binomial $p=2.4\times10^{-6}$). Such an asymmetry is not compatible with symmetric serving noise; the prompts were identical, so it points to a systematic difference between the two server sessions (for example batch composition or cache state) that we cannot identify from the logs. As a consequence, about half of the $+3.91$-point base versus self-correction difference for Qwen2.5-Coder-7B is not attributable to repair, whereas the within-run estimate ($+2.28$ points) is unaffected. Under symmetric noise, the observed flip counts imply 95\% half-widths of \halfBaseRange{} points for a difference in EX between two runs for the six base versus self-correction pairs of the Qwen2.5-Coder, Llama and OpenRouter runs (\halfAllRange{} over all their self-correction pairs, \clHalfRange{} for CodeLlama); because the flip rates are lower bounds, we treat differences below about one point as within serving noise.

\begin{table}[ht]
\centering
\caption{How self-correction works (configuration $\to$ + self-correction). Repaired: items for which at least one repair round was triggered. Within-run: repaired items whose final query is correct (their drafts did not execute), as a count and in points of EX with a database-clustered 95\% CI; this is the primary estimand. $+/-$: items gained/lost relative to the separate base run, split by whether a repair occurred; changes on unrepaired items reflect run-to-run variation. Non-exec.: final predictions that fail to execute, including items without a model response (one item in the CodeLlama-13B base $\to$ + corr.\ row). Upper block: from the base configuration; lower block: from a linked configuration. Tokens: mean prompt plus completion tokens per question; calls: mean model calls per question with self-correction.}
\label{tab:mech}
\footnotesize
\setlength{\tabcolsep}{3pt}
\begin{tabular}{lrrlrrrrr}
\toprule
 & & \multicolumn{2}{c}{Within-run} & \multicolumn{2}{c}{vs.\ base run, $+/-$} & & & \\
Model & Repaired & $n$ & Points [CI] & repaired & unrepaired & Non-exec. & Tokens & Calls \\
\midrule
\TabMechBody
\bottomrule
\end{tabular}
\end{table}

\paragraph{What the remaining errors are.} After self-correction, nearly all remaining failures are queries that execute but return a wrong result. For Qwen2.5-Coder-32B self-hosted, 640 of the 672 wrong items (95\%) execute, 24 do not, 5 are gold-failure items and 3 have no execution status because the gold query failed under load in the released scoring pass; for Llama-3.3-70B the counts are 658, 17 and 5 of 680 (97\%), and for the OpenRouter run 648, 24 and 5 of 677 (96\%). Execution status is taken from the harness record of each run and correctness from the official re-scoring. An execution-error signal therefore has little left to act on, and further gains would require semantic feedback rather than error messages. We did not categorize the semantic errors further.

\subsection*{RQ3: schema linking}

No schema-linking configuration significantly improved EX (Table~\ref{tab:sig}). Lexical linking significantly reduced EX for Llama-3.3-70B ($-4.24$ points, $p_{\mathrm{Holm}}=$\,\pHLlamaSeventyLink). The reductions for Qwen2.5-Coder-14B ($-2.15$ points, raw $p=0.0072$) and Llama-3.1-8B ($-2.35$ points, raw $p=0.0058$) are not significant after Holm correction ($p_{\mathrm{Holm}}=$\,\pHQwenFourteenLink{} and \pHLlamaEightLink); for Qwen2.5-Coder-7B and -32B and the OpenRouter run the differences ($+0.85$, $-1.17$ and $-1.04$ points) are not significant.

\begin{figure}[ht]
\centering
\includegraphics[width=0.8\linewidth]{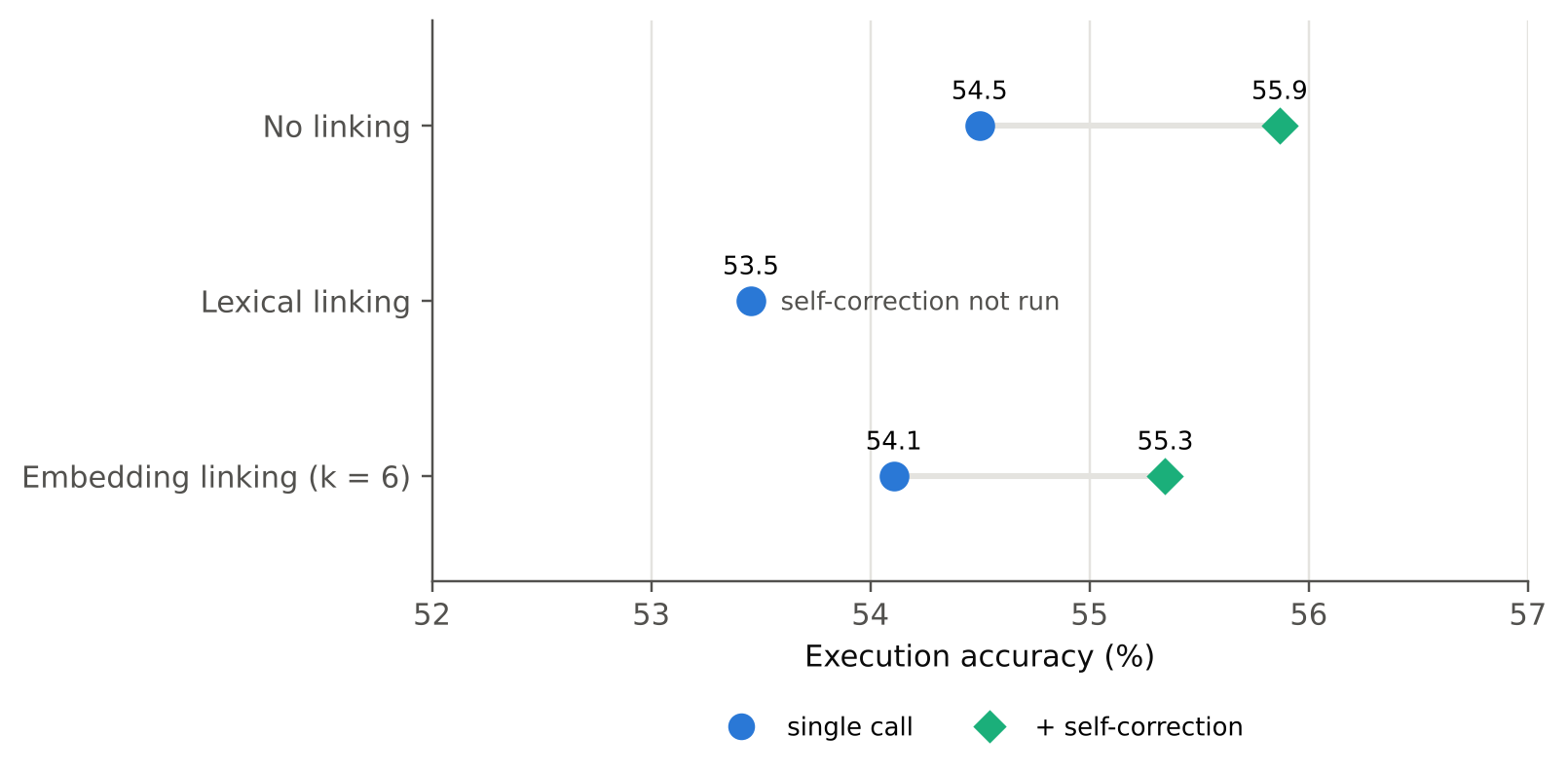}
\caption{Schema linking on the whitespace-repaired OpenRouter run of Qwen2.5-Coder-32B. Circles: single call; diamonds: with self-correction (not run for lexical linking). Neither linker differs significantly from no linking ($p=0.15$ lexical, $p=0.56$ embedding; Table~\ref{tab:sig}), although the embedding linker has higher gold-table recall (96.5\% versus 94.0\%).}
\label{fig:linking}
\end{figure}

A common objection is that a lexical pruner is too weak a baseline. On the OpenRouter run of Qwen2.5-Coder-32B we therefore compared the lexical linker with an embedding linker (Figure~\ref{fig:linking}). The embedding linker has higher mean gold-table recall (96.53\% versus 93.95\% over all questions; it retains every gold table for 91.66\% versus 87.22\% of questions). However, it keeps the full schema for the 652 questions on databases with at most six tables, and on the 882 questions where it does prune, its recall (94.10\%) is close to that of the lexical linker on the same questions (93.77\%). Its EX did not differ detectably from no linking ($-0.39$ points, $p=0.56$, clustered CI $-1.00$ to $0.25$) or from lexical linking ($+0.65$ points, $p=0.42$).
For CodeLlama (Table~\ref{tab:cl}), neither linker differed significantly from no linking after Holm correction: lexical linking changed EX by \clSevenLinkD{} (7B, raw $p=$\,\clSevenLinkP) and \clThirteenLinkD{} points (13B), and embedding linking by \clSevenEmbD{} (7B, $p_{\mathrm{Holm}}=$\,\clSevenEmbH) and \clThirteenEmbD{} points (13B, $p_{\mathrm{Holm}}=$\,\clThirteenEmbH). For CodeLlama-7B, however, the embedding linker was significantly worse than the lexical linker (\clSevenEmbLexD{} points, $p_{\mathrm{Holm}}=$\,\clSevenEmbLexH), despite its higher gold-table recall; for CodeLlama-13B the two linkers did not differ (\clThirteenEmbLexD{} points, $p_{\mathrm{Holm}}=$\,\clThirteenEmbLexH). Higher table recall therefore did not translate into higher accuracy for any model tested, and for the weakest model the choice of linker mattered more than whether linking was used at all.
The token saving from linking is modest (Table~\ref{tab:cost}): 421 to 355 mean tokens per question for lexical linking on Qwen2.5-Coder-32B and 421 to 393 for embedding linking on the OpenRouter run.

\begin{table}[ht]
\centering
\caption{Paired exact McNemar tests for the CodeLlama re-runs (self-hosted, fp16). $p_{\mathrm{Holm}}$ is adjusted over the full family of \holmM{} comparisons; CI: database-clustered bootstrap 95\% interval for $\Delta$.}
\label{tab:cl}
\footnotesize
\setlength{\tabcolsep}{4pt}
\begin{tabular}{llrrrrr}
\toprule
Model & Comparison & $\Delta$ & $+/-$ & $p$ & $p_{\mathrm{Holm}}$ & Clustered 95\% CI \\
\midrule
\TabCLBody
\bottomrule
\end{tabular}
\end{table}

\paragraph{Clustered sensitivity analysis.} Every comparison that is significant after Holm correction also has a database-clustered interval that excludes zero, so no confirmatory conclusion depends on treating questions as independent. The clustered intervals are, however, considerably wider than the McNemar-based intervals for comparisons whose effect varies between databases (for example lexical linking on Qwen2.5-Coder-14B, $-2.15$ points, CI $-5.61$ to $0.42$, and base versus self-correction for Qwen2.5-Coder-7B, CI $1.29$ to $8.17$ points). In the other direction, the clustered interval excludes zero for lexical linking on Llama-3.1-8B ($-2.35$ points, CI $-4.83$ to $-0.63$), and for embedding linking on CodeLlama-7B (\clSevenEmbD{} points, CI \clSevenEmbCI); neither is significant after Holm correction, and we report both as nominal reductions only.

\subsection*{RQ4: cost}

\begin{figure}[ht]
\centering
\includegraphics[width=0.85\linewidth]{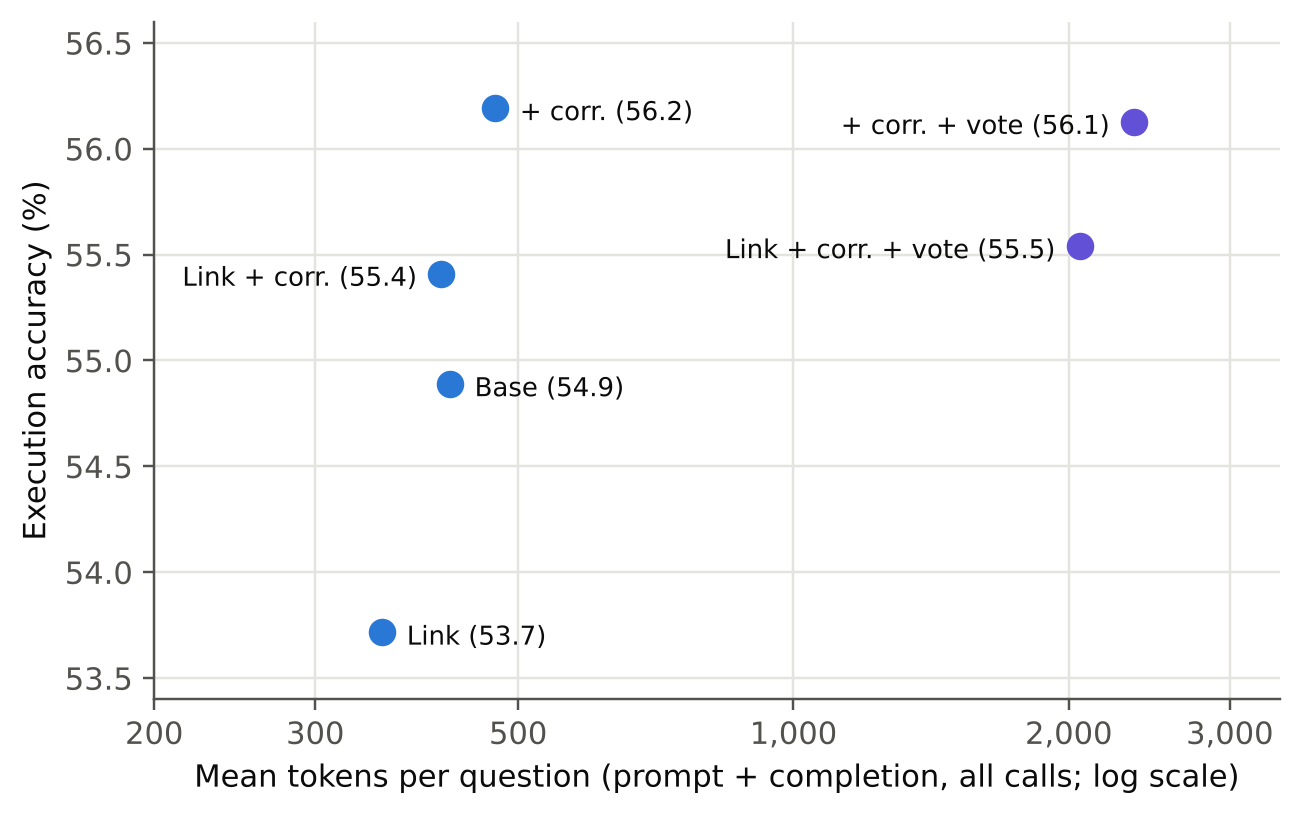}
\caption{Execution accuracy against mean tokens per question (log scale) for the Qwen2.5-Coder-32B self-hosted ablation. ``corr.'': self-correction; ``Link'': lexical schema linking; ``vote'': self-consistency with five samples (violet). Self-correction gives the largest gain per token; self-consistency adds about five times the tokens without a measurable gain.}
\label{fig:frontier}
\end{figure}

Table~\ref{tab:cost} reports tokens and model calls for every run. For the Qwen2.5-Coder, Llama and OpenRouter runs, self-correction costs 1.09 to 1.50 times the base tokens (1.06 to 1.40 model calls per question), because only drafts that fail to execute are repaired; for CodeLlama, whose drafts fail more often, self-correction raises the mean from \clSevenTokBase{} to \clSevenTokCorr{} tokens per question (7B, \clSevenTokRatio{} times) and from \clThirteenTokBase{} to \clThirteenTokCorr{} tokens (13B, \clThirteenTokRatio{} times). Self-consistency on Qwen2.5-Coder-32B costs 2358 instead of 472 tokens per question (4.99 times) and 5.48 model calls, raises the p95 summed model-call time per item from 10.6 to 50.0\,s at a concurrency of 64, and changes EX by $-0.07$ points ($p=1.00$; Figure~\ref{fig:frontier}). Lexical linking saves 16\% of tokens.

Wall-clock time is reported only where it was logged and is meaningful. For the API runs it reflects a shared, provider-managed service: for Llama-3.3-70B the self-correction run finished faster (348\,s) than the base run (499\,s) despite issuing more tokens, which illustrates that API timings are not a reliable cost measure; per-token API prices, which change over time, are a more appropriate basis there. We therefore report no dollar costs; tokens and model calls are the hardware-independent cost measures that transfer to a reader's own serving setup.

\begin{table}[ht]
\centering
\caption{Cost measures per configuration. Tokens: mean prompt plus completion tokens per question over all calls; calls: mean model calls per question; p95: 95th percentile of the summed model-call time per item under the stated concurrency, including client-side retries (not single-user latency); wall: total wall-clock time for 1534 questions where logged. CodeLlama p95 and wall-clock are not reported because the first pass included client timeouts and the timed-out items were completed in a separate retry pass.}
\label{tab:cost}
\footnotesize
\setlength{\tabcolsep}{4pt}
\begin{tabular}{llrrrr}
\toprule
Model (serving) & Configuration & Tokens & Calls & p95 (s) & Wall (s) \\
\midrule
\TabCostBody
\bottomrule
\end{tabular}
\end{table}

\section*{Discussion}

\paragraph{Self-correction as a default.} Self-correction was the only component with a consistent gain: its within-run effect was positive with a clustered interval excluding zero for every model, and the base versus self-correction comparison was significant after Holm correction for every model except Llama-3.3-70B, where run-to-run variation of the FP8 API masked it. Its cost is bounded because it acts only on drafts that fail to execute. Its mechanism also limits it: once nearly all final queries execute, the remaining errors are semantic, and error-message feedback cannot address them. This explains why the gain shrinks for the strongest models, whose drafts rarely fail (80 non-executing drafts for Llama-3.3-70B against 393 for Llama-3.1-8B).

\paragraph{Why linking does not help on BIRD.} Both linkers trade some recall for a shorter prompt. On BIRD, the full prompt is short (about 370 prompt tokens per question for the single-call configuration, including instructions, schema, question and evidence; repair rounds add more), five of the eleven databases have at most six tables, and the full schema fits easily into the context window of every model evaluated here. Dropping a needed table (which happens for 12.8\% of questions with the lexical linker and 8.3\% with the embedding linker) therefore costs more than the shorter prompt gains. This agrees with observations for stronger models \citep{deathlinking,longcontext}, whereas linking with recall safeguards has been reported to help in larger pipelines \citep{rslsql}. The conclusion is scoped to schemas of BIRD's size; for wide enterprise schemas such as those in Spider 2.0 \citep{spider2}, where the full schema may not fit or may distract the model, linking should be re-evaluated.

\paragraph{Serving effects.} Without the repair, the self-hosted and the OpenRouter runs of the same model differed by more than three points, which could easily have been attributed to the serving stack. The repaired run shows that most of this difference was an output defect of the single available provider, which removed spaces next to digits and made between 5\% and 6\% of the final queries unexecutable. Two lessons follow for practitioners. First, API outputs should be checked for systematic corruption before accuracy differences are attributed to precision or kernels; comparing against a self-hosted run of the same model, as here, makes such defects visible. Second, after the repair the cross-serving difference was below half a point, smaller than the run-to-run variation we observe within serving stacks, so we find no evidence that this provider's undisclosed serving precision materially changes Text-to-SQL accuracy for this model. This conclusion rests on one model and one provider.

\paragraph{Positioning against published numbers.} The protocol is deliberately minimal: one zero-shot call, no in-context examples and no fine-tuning. The Qwen2.5-Coder technical report gives 58.4 on BIRD for the 32B-Instruct model without stating the split or metric \citep{qwen25coder}; our 54.89\% under a single zero-shot prompt is in that neighbourhood, but the two figures are not directly comparable. Systems that add fine-tuning or reinforcement learning reach substantially higher development-set EX, for example Arctic-Text2SQL-R1 with 70.5\% at 32B and 70.1\% at 14B as reported by its authors \citep{arctic}. The contribution of this study is therefore not a competitive score but a controlled, paired comparison of inexpensive recipe components under documented serving. Because BIRD labels contain documented annotation errors \citep{birderrors} and run-to-run variation is of the order of one point, small absolute differences should be read with care.

\paragraph{Practical recommendation.} For private Text-to-SQL on schemas of BIRD's size, the evidence supports serving a strong, recent code model (in our data, at least 14B parameters, with diminishing and serving-confounded returns beyond that) and adding execution-guided self-correction. Schema linking did not help any model tested. Self-consistency did not help in the one model tested, and at about five times the tokens it would need a clear gain elsewhere to be justified. Accuracy should be compared only within a fixed serving stack and with paired tests, and API outputs should be checked for serving defects before conclusions about the stack are drawn.

\subsection*{Limitations}

(i) A single zero-shot prompt template is used; prompt sensitivity is not measured, and other prompts could change both levels and component effects. The schema serialization is minimal (unquoted column names containing spaces or parentheses, no types, keys or sample values), which lowers absolute EX and contributes to the non-executing queries. (ii) Only the BIRD development split is used; other benchmarks, and in particular wide-schema enterprise databases, may behave differently. (iii) BIRD contains annotation errors \citep{birderrors}, which add label noise to all EX values. (iv) Contamination cannot be excluded: the training data of the models may include BIRD questions or queries. A simple probe finds that 4.2--5.6\% of base-run predictions of the 2024 models (Qwen2.5-Coder, Llama-3.x and the OpenRouter run) match the gold query exactly after normalization, and \clSevenEM\% and \clThirteenEM\% for CodeLlama-7B and -13B. These rates are low and similar across the 2024 models, which argues against wholesale verbatim memorization, but they do not rule out subtler exposure, and the difference between generations is one more reason why the CodeLlama contrast in RQ1 is not a test of a generation effect. (v) The embedding linker's $k=6$ was chosen on the evaluation split, which, if anything, favours the embedding linker. (vi) Run-to-run variation changes the correctness of at least \flipQwenShort\% of items between greedy runs for Qwen2.5-Coder self-hosted (\clFlipRangeShort\% for CodeLlama) and up to 3.7\% on the FP8 API; these are lower bounds, and we did not replicate base runs to estimate the variance directly, except for the two unrepaired OpenRouter runs, which agreed within 0.2 points. Gold results were computed per scoring pass and not cached offline; under the uniform rule, \goldFailN{} questions are scored wrong in every run, so the maximum attainable EX is \goldMaxEX\%. (vii) Model size and serving stack change together between Qwen2.5-Coder-14B and -32B, so the size comparison is descriptive. (viii) The CodeLlama-34B run of the preprint was affected by the database-alteration defect and could not be re-run on the hardware available; it is not reported, and its released predictions are kept for transparency only. The CodeLlama re-runs needed a retry pass for requests that timed out on the client, and the 3072-token context limit of the T4 server rejected one repair round that followed a maximal-length draft. (ix) The Llama-3.x models were evaluated only through an FP8 API, and the precision of the OpenRouter serving is not disclosed by the provider; OpenRouter results rest on a single provider and on the whitespace repair. (x) The released Qwen2.5-Coder self-hosted runs were produced with an earlier harness revision (repair not triggered by timeouts, older extractor affecting 0--5 items per run, no wall-clock logging); they are re-scored but not regenerated. (xi) Self-consistency was run only for Qwen2.5-Coder-32B self-hosted, and only the selected candidate was stored.

\section*{Conclusions}

On the BIRD development split, open code models served by their operator reach about 55\% execution accuracy with a single zero-shot call at 32B. Within Qwen2.5-Coder, accuracy rises steeply from 7B to 14B and less from 14B to 32B, where the comparison is also confounded with the serving stack. Execution-guided self-correction is the only component with a consistent gain; it operates by repairing non-executing queries, adds about 1 to 4.5 points within a run and costs 1.1 to 1.5 times the tokens (about 1.8 to 1.9 times for CodeLlama, whose drafts fail more often). Schema linking, with either a lexical or a higher-recall embedding linker, yields no measurable gain, and five-sample self-consistency yielded none in the one model tested. An apparent three-point effect of the serving stack turned out to be mostly an output defect of one API provider; after repairing it, the same model through the API and self-hosted differed by less than half a point. Run-to-run variation is large enough that comparisons should be paired, made within one serving configuration, and interpreted with a margin of about one point. All reported numbers can be recomputed from the released predictions.

\section*{Data and code availability}

The harness, configuration files, prompt templates (\texttt{src/nl2sql\_bench/pipeline.py}), schema serializer and linkers, the whitespace repair (\texttt{src/nl2sql\_bench/textfix.py}), per-item predictions (\texttt{results/*.jsonl}) and per-run summaries are available on GitHub (\url{https://github.com/beskvladimir-create/nl2sql-onprem-bench}) and archived on Zenodo (release v0.3.0, DOI: 10.5281/zenodo.23445116) under the MIT license. The release used for this article contains the corrected harness, the per-item official re-scoring of all released runs, the OpenRouter runs (the whitespace-repaired run and the two unrepaired runs), the CodeLlama re-runs with their retry pass and merge script, and the analysis scripts (\texttt{paper\_numbers.py}, \texttt{make\_figures.py}) that regenerate every number, table and figure. The BIRD dataset is third-party and is obtained from its authors under their terms (\url{https://bird-bench.github.io/}); it is not redistributed.

\section*{Funding}

This research received no external funding. Cloud GPU rental and API access were paid for by the author; the CodeLlama re-runs used free Kaggle notebook GPUs.

\section*{Competing interests}

The author operates Besk Tech, which develops software products based on open-weight language models for closed-network deployments. Besk Tech did not fund this work, and none of its products is evaluated in this study.

\bibliography{references}

\end{document}